\documentclass{article}

\usepackage{PRIMEarxiv}

\usepackage[utf8]{inputenc} % allow utf-8 input
\usepackage[T1]{fontenc}    % use 8-bit T1 fonts
\usepackage{hyperref}       % hyperlinks
\usepackage{url}            % simple URL typesetting
\usepackage{booktabs}       % professional-quality tables
\usepackage{amsfonts}       % blackboard math symbols
\usepackage{nicefrac}       % compact symbols for 1/2, etc.
\usepackage{microtype}      % microtypography
\usepackage{lipsum}
\usepackage{fancyhdr}       % header
\usepackage{graphicx}       % graphics
\graphicspath{{media/}}     % organize your images and other figures under media/ folder

\title{Digitizing Coaching Intelligence: An Agentic Framework for Holistic Athlete Profiling using VLM and RAG
}

\author{
  Deep Ghosal, Ishani Sen, Wazib Ansar, Amlan Chakrabarti \\
  A.K. Choudhury School of Information Technology \\
  University of Calcutta \\
  Kolkata, India \\
  \texttt{\{deepghosal445, ishani.sen2001\}@gmail.com, \{waakcs\_rs, acakcs\}@caluniv.ac.in} \\
}

\begin{document}
\maketitle

\begin{abstract}
Athlete assessment is a critical process for tracking physical progress and identifying elite talent. However, during mass recruitment drives, traditional methods rely on manual observation, which is inherently subjective and unscalable, or basic computer vision (CV) systems limited to quantitative repetition counting. These standard approaches lack the ``coaching intelligence'' required to evaluate qualitative physiological markers such as form degradation, spinal articulation, and fatigue. This paper presents a novel, LLM-based hybrid agentic framework for automated, holistic athlete profiling that strictly aligns with the Sports Authority of India (SAI) assessment protocols. Orchestrated via \texttt{LangGraph}, our dual-pipeline architecture synthesizes the geometric precision of CV (\texttt{MediaPipe}) for kinematic tracking with the semantic reasoning of Vision-Language Models (\texttt{Llama-4-scout}). To overcome the latency and token constraints associated with multimodal video processing, we introduce a $3 \times 3$ ``Smart Grid'' temporal chunking strategy, reducing computational overhead by over 88\% while preserving critical temporal continuity. To ensure data integrity and mitigate hallucination, the framework pioneers an autonomous ``LLM-as-a-Judge'' self-correction loop that cross-references quantitative and qualitative metrics before persistence. Finally, we implement a dual-persistence Retrieval-Augmented Generation (RAG) pipeline utilizing a vector search engine (\texttt{ChromaDB}). This enables coaches to bypass rigid SQL databases and perform complex semantic queries (e.g., ``Identify athletes with high endurance but poor core rigidity'') using natural language. Experimental results demonstrate that this multi-agent approach significantly bridges the gap between raw biometric tracking and actionable coaching insights, offering a scalable, objective solution for national talent identification.
\end{abstract}

% keywords can be removed

\section{Introduction}

\subsection{Problem Definition \& Its Significance}
In the landscape of national sports development, particularly within populous nations like India, the process of talent identification and performance assessment faces a critical bottleneck: scalability versus depth. Organizations such as the Sports Authority of India (SAI) conduct mass recruitment drives where thousands of candidates are evaluated simultaneously. The current methodology relies heavily on manual observation by human coaches. This traditional approach is inherently subjective, labor-intensive, and prone to ``observer fatigue,'' leading to inconsistencies where elite talent might be overlooked due to human error or bias.

While recent technological advancements have introduced Computer Vision (CV) into sports analytics, existing solutions remain fundamentally limited. Standard CV systems are purely quantitative; they function as digital counters, tracking repetitions (e.g., number of push-ups) without understanding the quality of the movement. They fail to detect crucial physiological markers such as form degradation, onset of fatigue, or compensatory movements---nuances that a human coach intuitively grasps. A candidate performing 50 repetitions with poor form may statistically outrank a candidate performing 40 repetitions with perfect biomechanics, resulting in flawed profiling. The core problem, therefore, is the lack of an automated system that possesses ``coaching intelligence''---the ability to synthesize geometric data with qualitative reasoning.

The significance of this research lies in its proposal of an LLM-based Agentic Framework to bridge this gap. By moving beyond static algorithms to a multi-agent system, we enable a holistic assessment that mirrors expert human analysis. This system integrates the precision of computer vision with the semantic understanding of Vision-Language Models (VLMs). For the Sports Authority of India and athletic assessors, this innovation is transformative. It allows for the democratization of high-quality coaching feedback; a candidate in a remote district can receive the same level of analytical scrutiny as one in a national camp. Furthermore, the implementation of Retrieval-Augmented Generation (RAG) \cite{19, 23} allows assessors to interact with vast datasets using natural language (e.g., ``Show me athletes with high endurance but poor core stability''), turning raw data into actionable insights instantly.

\subsection{Gaps in Existing Research}
Recent advancements in automated fitness monitoring have largely transitioned from wearable sensors to non-intrusive computer vision techniques. Studies like Fitcam \cite{1} and Shi et al. \cite{6} demonstrate the efficacy of deep learning (OpenPose, CPM) for counting standard exercises like push-ups and sit-ups with high accuracy. Other approaches, such as Park and Baek \cite{2} and Wang and Kong \cite{4}, employ parametric rules and KNN algorithms to classify specific movement phases and geometric forms. 

While frameworks like Ferreira et al. \cite{5} attempt validation using state machines, these existing solutions remain fundamentally limited to quantitative metrics. They function primarily as digital counters, excelling at tracking volume but failing to capture the quality of movement comprehensively. They lack the semantic understanding to detect ``soft biometrics'' like fatigue, struggle, or temporal form degradation, often missing the nuanced insights required for elite athletic profiling \cite{7}.

\subsection{Addressing the Research Gap}
This research addresses the gap between simple repetition counting and holistic athletic profiling by proposing a multi-layered \textit{Hybrid Agentic Framework} strictly aligned with the Sports Authority of India (SAI) and Khelo India assessment protocols. We move beyond standard computer vision by integrating Vision-Language Models (VLMs) directly into the analysis pipeline to evaluate complex biomechanical markers, specifically \textbf{Spinal Articulation} and \textbf{Core Rigidity}. 

While our dedicated CV agent handles geometric precision and counting accuracy found in prior works \cite{2,6}, our novel VLM agent introduces ``Coaching Intelligence.'' To overcome the high latency typically associated with multimodal models, we implemented a $3 \times 3$ \textbf{``Smart Grid'' Temporal Chunking Strategy}, reducing computational overhead by over 88\% while successfully capturing qualitative markers like shivering, grimacing, and tempo changes. 

A critical innovation in our framework is the multimodal self-correction loop powered by an ``LLM-as-a-Judge'' (Agent 4). This agent actively cross-references quantitative CV metrics with qualitative VLM insights to ensure logical consistency, triggering a re-evaluation if contradictions occur. Furthermore, we implement a hybrid data layer---syncing Google Sheets for structured relational data and ChromaDB for semantic vector embeddings. This powers a Retrieval-Augmented Generation (RAG) pipeline that transforms rigid data logging into natural language insights. This allows coaches to query complex performance variables (e.g., ``Find athletes with good stamina but poor core stability''), effectively democratizing elite-level sports analytics for scalable, mass-population testing.

\subsection{Our Contributions}

\subsubsection{Development of an SAI-Compliant Hybrid Agentic Framework}
We introduce a novel multi-agent architecture orchestrated via \texttt{LangGraph} that seamlessly integrates diverse AI modalities. By decoupling video processing into specialized sub-agents---one for Vision-Language Model (VLM)-based qualitative analysis (assessing fatigue, \textit{Spinal Articulation}, and \textit{Core Rigidity}) and another for Computer Vision (CV)-based quantitative tracking (repetition counts, angular kinematics)---we achieve a holistic assessment. This framework significantly surpasses standalone computer vision by strictly aligning its deterministic output with the official Sports Authority of India (SAI) and Khelo India assessment protocols.

\subsubsection{Digitization of ``Coaching Intelligence''}
Unlike traditional trackers that output raw, isolated metrics, our system features a dedicated ``Assessor Agent'' (Agent 3) that synthesizes multi-modal inputs into a coherent profile. This agent applies a weighted scoring algorithm---balancing repetition volume, biomechanical form adherence, and visual endurance---to generate an actionable ``Athlete Score'' alongside professional natural language feedback. This effectively replicates the subjective expertise of a human coach for mass-scale application while enforcing a strict, standardized grading rubric.

\begin{table}[h!]
\centering
\caption{Athlete to Assessor Comparison}
\label{table:comparison}
\begin{tabular}{|p{3.5cm}|p{4.5cm}|p{4.5cm}|}
\hline
\textbf{Parameter} & \textbf{Manual Assessment (Traditional)} & \textbf{Proposed AI-Agentic Framework} \\ \hline
Athlete-to-Assessor Ratio & 50:1 to 100:1 (during mass trials) & Continuous 1:1 digital monitoring \\ \hline
Assessment Time & 15--20 minutes (full battery) & Real-time (during activity) \\ \hline
Data Throughput & Limited visible metrics & High-frequency biomechanical data \\ \hline
Error Margin & High (Fatigue, Bias, Subjectivity) & Low (Deterministic CV + LLM Logic) \\ \hline
Qualitative Depth & Relies on coach's notes/memory & Immediate semantic feedback \\ \hline
\end{tabular}
\newline
Source: \href{https://www.pib.gov.in/PressReleasePage.aspx?PRID=1556774&reg=3&lang=2#:~:text=Identification%20of%20sporting%20talent%20will,at%20the%20rate%20of%20Rs.}{Press Information Bureau}
\end{table}

\subsubsection{Multimodal Self-Correction and Quality Assurance}
To eliminate model hallucination and ensure data integrity, we introduce an autonomous ``LLM-as-a-Judge'' (Agent 4). This novel quality control agent actively cross-references the quantitative geometric data from the CV module with the qualitative insights from the VLM. If logical contradictions are detected (e.g., CV reports zero valid reps but VLM praises perfect form), the system triggers a dynamic, low-confidence re-inference loop. This guarantees that only highly confident, logically verified assessments proceed to the database\cite{11}.

\subsubsection{Optimized `Smart Grid' VLM Processing for Temporal Analysis}
To address the token limitations, high latency, and computational costs inherent in processing long video inputs with Large Language Models, we engineered a robust $3 \times 3$ \textbf{``Smart Grid'' temporal sampling strategy}. This method condenses sequential video frames into composite grid images, reducing computational overhead by over 88\%. This ensures that critical temporal markers of fatigue and form degradation are accurately captured across the entire exercise session without suffering from data loss or API constraints.

\subsubsection{Hybrid Persistence and Semantic Retrieval for Actionable Insights}
We address the data accessibility bottleneck by implementing a dual-persistence strategy coupled with a Retrieval-Augmented Generation (RAG) pipeline tailored for sports analytics. Structured quantitative data is synced to relational spreadsheets, while semantic assessor feedback is embedded into a vector search engine (\texttt{ChromaDB}). This allows assessors to query massive athlete datasets using natural language (e.g., ``Find candidates with high rep counts but poor core rigidity''), instantly converting raw performance logs into strategic recruitment insights without needing SQL expertise.

\section{Literature Survey}

The field of automated fitness monitoring and athletic assessment has evolved rapidly, transitioning from wearable-based tracking to sophisticated non-intrusive computer vision systems. This section reviews the core methodologies and recent advancements that inform our hybrid agentic framework.

The 2024 ``Fitcam'' study \cite{1} introduced an automated deep learning system designed to monitor unsupervised fitness routines and prevent injuries. Targeting common exercises such as push-ups, sit-ups, and squats, the methodology utilizes a multi-stage pipeline combining the \texttt{OpenPose} library for skeletal keypoint extraction with \textit{Long Short-Term Memory} (LSTM) networks to analyze temporal movement patterns. A standout feature is its gesture-based activation, which provides a scalable, vision-based solution for home fitness environments.

Addressing the specific requirements of military and athletic assessments, Park and Baek \cite{2} proposed a real-time push-up monitoring system. Using \texttt{OpenPose} to extract 15 skeletal keypoints, the system identifies ``Up,'' ``Down,'' and ``Starting'' positions through spatial-temporal analysis. Unlike purely data-driven models, this approach employs ``parametric'' classification---applying specific geometric rules, such as back alignment and chest depth, to distinguish between correct and incorrect form, even in advanced variations.

Research presented at the 2022 NAFOSTED Conference \cite{3} explored the concept of a virtual ``personal trainer'' to address the surge in home workouts. Their system integrates deep learning with signal processing to analyze video data and track body joints in real-time, focusing on human keypoint detection to recognize specific movement phases while simultaneously tracking repetition volume and quality.

Advancements in computational efficiency are highlighted by Wang and Kong \cite{4}, who proposed a robust vision-based framework utilizing a specialized \texttt{YOLOv5s} target detector integrated with an optimized 33-keypoint \texttt{MediaPipe} Pose model. To classify movement phases, the system employs a \textit{K-Nearest Neighbors} (KNN) algorithm that assesses similarity between current feature vectors and training samples, achieving an impressive inference speed of 30 FPS and a counting accuracy of 99.5\%.The field of automated fitness monitoring and athletic assessment has evolved rapidly, building upon foundational vision-based tracking algorithms \cite{16} and transitioning to sophisticated non-intrusive skeletal estimation systems.

For complex training scenarios like CrossFit, Ferreira et al. \cite{5} proposed a system for real-time repetition counting and validation using 2D human pose estimation. Their architecture utilizes a \textit{Multi-Layer Perceptron} (MLP) integrated with an Encoder Network and employs a key pose augmentation strategy to overcome dataset imbalance, resulting in over 92\% precision for valid repetitions.

Specialized movement analysis, such as sit-up counting, has seen architectural refinement through the work of Shi et al. \cite{6}. They optimized a \textit{Convolutional Pose Machine} (CPM) by simplifying its architecture from six stages to four and introducing skip connections. This refinement significantly reduced model complexity and increased inference speed from 36.5 to 63.6 FPS while maintaining a 98.57\% counting accuracy.

Moving beyond individual tracking algorithms, Weakley et al. \cite{7} provided a comprehensive narrative review of the standards required for effective athletic assessment. They emphasize that while profiling is essential for talent identification, results depend heavily on a scientific framework of validity, reliability, and sensitivity. Their recommendations include prioritizing tests with high translational validity and utilizing advanced data visualization to transform raw metrics into actionable coaching insights.

Finally, a review by Khanal et al. \cite{8} analyzed the broader potential of non-contact computer vision technologies for physiological monitoring. By categorizing techniques like RGB-Depth cameras and thermal imaging, the authors demonstrate how computer vision offers a non-intrusive and cost-effective alternative to wearable sensors for tracking heart rate and muscle fatigue. This highlights the utility of vision-based monitoring in environments where minimizing physical contact improves user safety and compliance.

\section{Proposed Methodology}

This study proposes a novel, multi-agent framework orchestrated via \texttt{LangGraph}, designed to automate and standardize holistic athlete profiling in strict alignment with the Sports Authority of India (SAI) assessment protocols. The overarching architecture is structurally divided into two distinct yet highly interconnected pipelines: a Multimodal Data Ingestion Pipeline and a Semantic Retrieval Pipeline.

The ingestion phase synthesizes the deterministic, geometric precision of Computer Vision (CV) with the semantic reasoning capabilities of Vision-Language Models (VLMs). To overcome the high latency and computational constraints typically associated with multimodal video processing, we introduce a novel $3 \times 3$ ``Smart Grid'' temporal sampling strategy. Furthermore, the framework pioneers an autonomous ``LLM-as-a-Judge'' self-correction loop, ensuring that all cross-modal data is logically verified for integrity before being committed to a dual-persistence hybrid storage layer.

The secondary phase leverages a Retrieval-Augmented Generation (RAG) architecture to democratize data accessibility. By translating rigid relational data and dense vector embeddings into natural language, the system enables sports assessors and coaches to query complex performance variables dynamically. Together, these pipelines form a comprehensive, highly scalable solution for objective, high-throughput athletic talent identification.

\begin{figure}[htbp]
    \centering
    \includegraphics[width=0.8\textwidth]{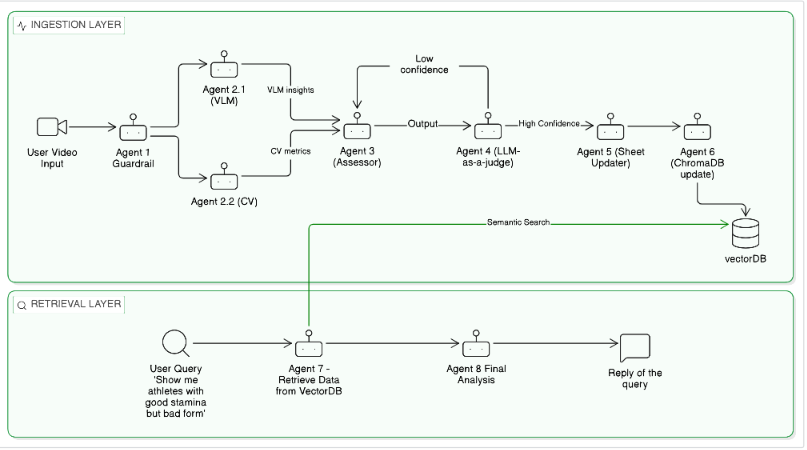}
    \caption{Work flow of the system}
    \label{fig:elbow_trajectory}
\end{figure}

\subsection{Multimodal Data Ingestion Pipeline}
The ingestion phase serves as the foundation of the proposed framework, transforming unstructured raw video input into structured, quantifiable, and SAI-compliant athlete profiles. This process is orchestrated by a specialized chain of autonomous agents designed to handle specific modalities of data---temporal, geometric, and semantic. By decoupling the processing logic, the system ensures that high-fidelity quantitative metrics are captured alongside nuanced qualitative observations without suffering from computational bottlenecks or model hallucination. 

\begin{figure}[htbp]
    \centering
    \includegraphics[width=0.8\textwidth]{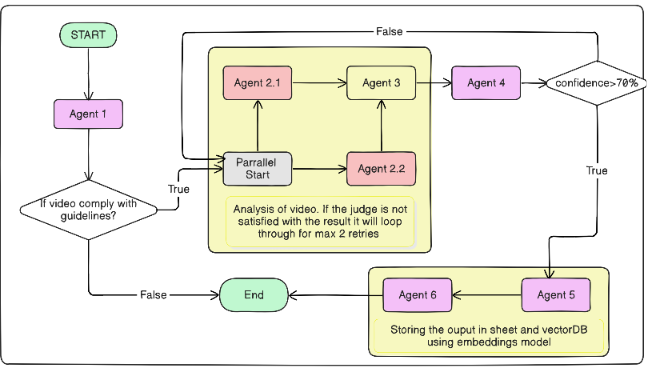}
    \caption{Agentic workflow of the ingestion part}
\end{figure}

\subsubsection{Data Acquisition \& Guardrail Mechanism}
Before deep analysis begins, raw video footage must be validated to prevent the system from processing corrupted, irrelevant, or malicious data. This is handled by Agent 1, the \textit{Guardrail Mechanism}\cite{12}\cite{13}.

Upon receiving a video, Agent 1 performs a rapid preliminary scan using a lightweight object-detection model. Its primary function is twofold:
\begin{itemize}
    \item \textbf{Exercise Verification:} It confirms that the video actually contains an athlete performing one of the recognizable SAI protocols (e.g., Push-ups or Sit-ups).
    \item \textbf{Content Assurance:} It checks for critical anomalies, such as severe camera occlusion, the absence of a human subject, or frame corruption.
\end{itemize}
If an anomaly is detected, Agent 1 halts the pipeline and logs a rejection error, preventing downstream agents from wasting token computations. If the video passes validation, the data is passed to the acquisition phase\cite{14}.

\subsubsection{Data Acquisition \& ``Smart Grid'' Temporal Chunking (\texorpdfstring{$3 \times 3$}{3x3} Optimization)} 
Processing high-definition video directly through Large Multimodal Models (LMMs) presents significant challenges regarding API token consumption, latency, and context window limitations. A naive approach of feeding all frames simultaneously often leads to ``loss in the middle'' phenomena, where the model hallucinates details or fails to track temporal progression accurately over long workout sessions.

To mitigate this, we engineered a robust ``Smart Grid'' Temporal Chunking Strategy. The video is first sampled at a target frame rate (e.g., 2 FPS) to reduce redundancy while preserving essential motion fidelity. Instead of sending these frames individually, the system condenses them into composite grid images.

Specifically, we utilize a $3 \times 3$ Grid Matrix. For a sequence of 9 continuous frames, the system stitches them together into a single, high-resolution image file. Each grid is read strictly from Left-to-Right, Top-to-Bottom (Row 1: Frames 1--3, Row 2: Frames 4--6, Row 3: Frames 7--9). Furthermore, a prominent green timestamp (e.g., ``T: 2.5s'') is digitally watermarked onto the bottom-left corner of each sub-frame.

This $3 \times 3$ Chunking Strategy provides critical advantages:
\begin{itemize}
    \item \textbf{Contextual Continuity:} It allows the Vision-Language Model to observe a continuous ``story'' of movement within a single API call, recognizing micro-movements like the trembling of arms during a push-up descent.
    \item \textbf{Massive Token Efficiency:} By consolidating 9 frames into 1 image, we reduce the token load and API requests by approximately 88\%.
    \item \textbf{Temporal Anchoring:} The embedded timestamps force the VLM to temporally anchor its observations, enabling it to pinpoint exactly when a form breakdown occurred (e.g., ``Hips sagged at T: 8.0s'').
\end{itemize}

\subsubsection{Qualitative Biomechanical Analysis (Agent 2.1: Vision-Language Model)}
Agent 2.1 is responsible for ``Coaching Intuition''---the qualitative assessment of physical exertion, form stability, and adherence to SAI guidelines. This agent processes the $3 \times 3$ Smart Grids generated in the acquisition phase using an advanced Vision-Language Model (such as \texttt{Llama-3.2-Vision}).

Unlike standard CV classifiers that track specific joint coordinates, Agent 2.1 is explicitly prompted with the rigid SAI Assessment Rules. It is tasked with evaluating soft biometrics and holistic form variables, specifically focusing on:
\begin{itemize}
    \item \textbf{Spinal Articulation (Sit-ups):} Evaluating whether the athlete curls their spine smoothly or uses jerky, momentum-driven neck movements to rise.
    \item \textbf{Core Rigidity (Push-ups):} Assessing if the athlete maintains a perfectly straight body alignment or suffers from ``hip sagging'' or ``piking'' under fatigue.
    \item \textbf{Visual Stamina Analysis:} Detecting physical struggles, such as extended resting periods on the floor, facial grimacing, or trembling limbs.
\end{itemize}

For each video, Agent 2.1 outputs a structured JSON object containing a global Endurance Score (0--100) and specific, timestamped qualitative feedback, adding a layer of depth that distinguishes between effortless performance and performance marred by severe form breakdown.

\subsubsection{Quantitative Kinematic Analysis (Agent 2.2: Computer Vision)} 
Parallel to the VLM analysis, Agent 2.2 executes a strictly quantitative assessment using classical Computer Vision. This agent utilizes the \texttt{MediaPipe} Pose library to extract 33 distinct 3D skeletal landmarks from the raw video frames. Its primary objective is to rigorously count valid repetitions and calculate geometric limits.

For kinematic analysis, the agent calculates joint angles (e.g., $\theta$) using the vector dot product formula. Given three landmarks---shoulder ($A$), hip ($B$), and knee ($C$)---the angle is computed to ensure the athlete is reaching the required depth or maintaining specific postures.

A robust Finite State Machine (FSM), integrated with a hysteresis loop, is employed to track the repetition cycle. For instance, a repetition is only incremented when the calculated angle crosses strict, pre-defined thresholds denoting the ``UP'' state, crosses a secondary threshold denoting the ``DOWN'' state, and successfully returns to the ``UP'' state without early termination (which would be flagged as a partial or half-repetition). Agent 2.2 outputs these hard numerical values (e.g., Valid Reps = 45, Average Depth = $92.4^\circ$) to be merged with the qualitative data downstream
.

\subsection{Synthesis, Quality Assurance, and Persistence}
Following the parallel processing of quantitative kinematics (Agent 2.2) and qualitative biomechanics (Agent 2.1), the system enters the synthesis phase. This stage is critical for bridging the gap between raw, multi-modal data streams and actionable, human-readable insights. It mimics the cognitive process of a seasoned coach who weighs various performance factors to form a holistic opinion of an athlete, while introducing a novel layer of automated quality control to ensure data integrity before persistence.

\subsubsection{Aggregation \& SAI-Compliant Scoring (Agent 3: Assessor)}
Agent 3 serves as the central intelligence node of the ingestion pipeline, functioning as an aggregation and scoring engine. It receives two distinct data payloads: the geometric metrics from the CV module (valid repetition counts, depth angles, and fault logs like ``Elbow Flare'') and the physiological observations from the VLM (Endurance Score, spinal articulation, core rigidity, and signs of struggle).

The core innovation of Agent 3 lies in its SAI-Compliant Weighted Scoring Algorithm. Unlike traditional trackers that rank athletes solely by the number of repetitions completed, our system computes a composite Athlete Score ($S_{final}$). This score reflects a balanced view of performance, ensuring that an athlete who performs 50 poor-form push-ups does not outrank one who performs 40 high-quality ones.

The final score is calculated using a linear weighted sum equation:
$$S_{final}=(W_1 \times Reps)+(W_2 \times F_{qual})+(W_3 \times E_{vlm})$$

Where:
\begin{itemize}
    \item $Reps$ is the normalized repetition count (scaled against an age/gender benchmark).
    \item $F_{qual}$ is the Form Quality score, derived by deducting penalty points for every detected biomechanical fault (e.g., deducting points for ``Hips Sagging'' detected by the CV agent or ``Poor Core Rigidity'' flagged by the VLM).
    \item $E_{vlm}$ is the visual endurance score provided by Agent 2.1.
    \item $W_1, W_2, W_3$ are configurable weights (e.g., 0.4, 0.4, 0.2), emphasizing that biomechanical technique is as valuable as repetition volume.
\end{itemize}

Beyond numerical scoring, Agent 3 utilizes a Large Language Model (e.g., Llama 3) to generate a Natural Language Feedback Report. By synthesizing the aggregated metrics, the agent produces a professional summary highlighting strengths and specific areas for improvement, providing the athlete with actionable guidance.

\subsubsection{Multimodal Self-Correction Loop (Agent 4: LLM-as-a-Judge)}
To eliminate model hallucination and ensure strict data integrity, we introduce an autonomous quality assurance mechanism: Agent 4 (LLM-as-a-Judge). Acting as an independent auditor, Agent 4 reviews the finalized report generated by the Assessor (Agent 3) and explicitly cross-references the quantitative CV data against the qualitative VLM insights to detect logical contradictions. For example, if the CV agent (Agent 2.2) logs zero valid repetitions because the athlete failed to reach a 90-degree depth, but the VLM agent (Agent 2.1) outputs praise for ``excellent form and stamina,'' Agent 4 catches this semantic mismatch.

To strengthen this quality control stage and prevent the system from incorrectly approving weak or contradictory assessments, the Judge agent utilizes a structured tool-calling approach rather than traditional prompt-only JSON parsing. This structural validation strictly enforces a predefined response schema, requiring the LLM to output specific fields: a Confidence Score (0--100\%), a justification reason, optional contradiction tags, and a severity level.

Upon evaluating the assessment, Agent 4 dictates the flow of the pipeline based on this output:
\begin{itemize}
    \item \textbf{High Confidence ($\ge 70\%$):} The data streams align logically. The validated report is approved and proceeds to the persistence layer.
    \item \textbf{Low Confidence ($< 70\%$):} A logical contradiction is detected. Agent 4 flags the specific error (e.g., Contradiction: High VLM praise vs. Zero CV reps) and triggers a dynamic Re-inference Loop. The data is routed back to Agent 3---with the Judge's correction instructions included in the retry prompt---forcing the Assessor to re-evaluate the conflicting parameters.
\end{itemize}

Furthermore, to maximize pipeline safety, a ``fail-closed'' policy is embedded within this loop. If the Judge agent fails to produce a valid structured response (e.g., due to an API timeout or parsing failure), the system automatically assigns a low confidence score. It then triggers the re-analysis loop and flags the assessment for manual review if the maximum retry limit is exceeded.

Overall, the integration of structured tool-calling and fail-safe routing significantly improves the robustness, traceability, and trustworthiness of the final athlete assessment process. This self-correction loop guarantees that only highly confident, logically verified, and strictly SAI-compliant assessments are permanently recorded in the database.

\subsubsection{Hybrid Data Storage Strategy (Agent 5 \& Agent 6)}
Once the assessment passes the Judge's audit, the data must be securely stored. To support both traditional tabular reporting and advanced natural language querying, we implemented a dual-persistence Hybrid Data Storage Strategy.

\begin{itemize}
    \item \textbf{Relational Storage (Agent 5 - Sheet Updater):} Agent 5 acts as the system's structured database connector. It flattens the quantitative metrics---such as Timestamp, Athlete ID, Final Score, Valid Reps, and Average Depth Angle---and appends them as a new row in a cloud-based spreadsheet (e.g., Google Sheets). This provides coaches with a familiar, tabular ledger for tracking longitudinal progress and generating standard statistical reports.
    \item \textbf{Semantic Vector Storage (Agent 6 - ChromaDB Updater):} Parallel to Agent 5, Agent 6 handles the unstructured, qualitative data. It takes the rich, natural language feedback generated by the Assessor (including notes on spinal articulation, core rigidity, and fatigue) and converts it into high-dimensional vector embeddings using an embedding model (e.g., \texttt{gemini/embeddings-001}). These vectors are then inserted into \texttt{ChromaDB}, an open-source vector database.
\end{itemize}

By storing the unstructured feedback as mathematical vectors, Agent 6 creates a searchable ``semantic map'' of the entire athlete population. This hybrid approach---combining the exactness of relational tables with the semantic depth of a vector store---is the foundational requirement for the downstream Retrieval-Augmented Generation (RAG) pipeline.

\subsection{Semantic Retrieval \& Analysis (RAG Pipeline)}
The second half of the proposed framework addresses the critical challenge of data accessibility by implementing a Retrieval-Augmented Generation (RAG) pipeline. In traditional sports analytics, data gathered from athletes is often siloed in rigid relational databases, accessible only through complex SQL queries that demand exact keyword matches. This creates a significant bottleneck where coaches and assessors must rely on technical data analysts to extract meaningful insights\cite{18}.

The Retrieval Layer democratizes this process by allowing stakeholders to interact with the database using natural language queries, such as ``Show me athletes with good stamina but bad form.'' This pipeline transforms the passive data repository into an active, conversational analytical partner through two specialized agents, effectively shifting the paradigm from rigid ``Keyword Search'' to dynamic ``Concept Search.''\cite{15}

\begin{figure}[htbp]
    \centering
    \includegraphics[width=0.8\textwidth]{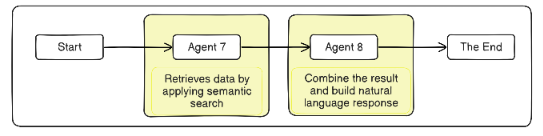}
    \caption{Agentic workflow of the Retreival part}
\end{figure}

\subsubsection{Context-Aware Vector Search (Agent 7: Retriever)}
Agent 7 serves as the entry point and search engine for the retrieval system. When a coach submits a natural language query, Agent 7 executes a semantic similarity search \cite{22} against the vector database (populated earlier by Agent 6 during the ingestion phase).

Rather than relying on exact phrasing---which is brittle and prone to vocabulary mismatch---this agent looks for semantic proximity. For instance, a query for ``weak core stability'' will mathematically map to vectors representing ``Hips Sagging'' or ``lumbar hyperextension,'' even if those exact keywords are not explicitly typed by the user. Programmatically, Agent 7 executes a \texttt{similarity\_search\_with\_score} function to isolate the top three ($k=3$) most relevant athlete documents matching the user's intent. The agent then parses these results into a structured \texttt{retrieved\_context} payload, which includes:

\begin{itemize}
    \item The raw text content of the athlete's assessment report.
    \item The associated metadata (such as athlete IDs, exercise types, scores, and timestamps).
    \item The mathematical \texttt{similarity\_score} to quantify the relevance of the match.
\end{itemize}

This robust context extraction ensures that no relevant performance data is overlooked simply due to a phrasing discrepancy.

\subsubsection{Strategic Insight Generation (Agent 8: Analyst)}
While Agent 7 locates the relevant data, Agent 8 (The Analyst) serves as the reasoning layer responsible for generating the final strategic insight. This agent utilizes a high-parameter Large Language Model configured with an exceptionally low temperature setting (0.1) via the Groq client. This hyper-parameter tuning is critical for minimizing model hallucination, ensuring that the generated responses remain strictly deterministic and factual.

Agent 8 is explicitly prompted to assume the persona of a ``Senior Head Coach and Sports Data Analyst.'' It ingests the user's original query alongside the JSON-formatted RAG context retrieved by Agent 7. To ensure the output is highly actionable and serves as an effective example of Explainable AI, Agent 8 operates under strict generation rules:

\begin{itemize}
    \item It must answer the user's question directly.
    \item It must explicitly cite specific data points from the retrieved context (e.g., citing a specific athlete's score or fault count).
    \item It must compare candidates when relevant to the query.
    \item It must conclude with a professional, data-driven recommendation.
\end{itemize}

The final output is constrained to an authoritative tone of under 200 words, ensuring that coaches receive concise, immediate, and easily digestible insights. Furthermore, to maintain system integrity, if Agent 7 fails to find relevant semantic matches, Agent 8 is programmed with a fallback response to gracefully inform the user that no athletes matching the specific criteria could be found in the database.

\section{Dataset and Experimental Setup}
To evaluate the efficacy of the proposed Agentic Framework, we conducted a series of experiments utilizing a curated multimodal dataset and a cloud-based inference environment. This section details the characteristics of the data used for validation and the specific software and hardware configurations employed to orchestrate the multi-agent workflow.

\subsection{Dataset Description}
Due to the lack of publicly available datasets containing standardized, high-quality recordings of both push-ups and sit-ups suitable for complex biomechanical profiling, we constructed a custom curated athletic dataset tailored for this research.

\subsubsection{Data Source \& Composition} 
The dataset comprises video samples collected from diverse environments to simulate real-world field testing conditions. Samples were sourced primarily from the \textbf{Kaggle LSTM Exercise Classification} dataset and the \textbf{Kinetics Human Action Video Dataset}, ensuring significant variability in lighting, background clutter, and camera angles. 

The dataset consists of 80 total video samples, categorized into two primary exercise classes:
\begin{itemize}
    \item \textbf{Push-ups (60 videos):} Focused on upper body strength, measuring elbow extension and core stability.
    \item \textbf{Sit-ups (60 videos):} Focused on abdominal endurance, measuring torso elevation and hip flexion.
\end{itemize}

\subsubsection{Experimental Matrix \& Data Points} 
To comprehensively evaluate the Hybrid Agentic Framework, each of the  videos was processed across an experimental matrix testing two distinct Vision-Language Models (VLMs) against two temporal 120 sampling rates ($2\,FPS$ and $3\,FPS$). This cross-validation approach generated a robust set of 480 unique data points for the final performance analysis.

\subsubsection{Video Characteristics} 
To test the real-world robustness of the Computer Vision agent, input data was deliberately not pre-processed for standardization:
\begin{itemize}
    \item \textbf{Resolution:} Varied from $480p$ to $720p$.
    \item \textbf{Frame Rate:} Original footage ranged from $30$ to $60\,FPS$. During the Ingestion Pipeline, these were down-sampled to target frequencies of $2\,FPS$ and $3\,FPS$.
    \item \textbf{Perspective:} Includes monocular views, primarily sagittal (side profile) and oblique angles, matching standard vantage points for official assessments.
\end{itemize}

\subsection{Experimental Setup \& Implementation Details}
The implementation was executed within a local \textit{Visual Studio Code} environment utilizing a \texttt{Python 3.12.4} runtime. Transitioning from cloud-notebook prototypes to a modular local environment allowed for strict version control and optimized environment variable management via \texttt{python-dotenv}.

\subsubsection{Agentic Orchestration and State Management}
The system backbone is \texttt{langgraph} (v1.1.8). Unlike linear chains, \texttt{LangGraph} allowed for the definition of the workflow as cyclical graph structures (\texttt{StateGraph}), essential for the Judge agent's dynamic retry loops. We defined a global state schema using Python’s \texttt{TypedDict} (\texttt{IngestionState} and \texttt{RetrievalState}) to act as shared memory for tracking variables like \texttt{cv\_rep\_count} and \texttt{judge\_feedback}.

\subsubsection{Computer Vision and Geometric Analysis}
For quantitative analysis (Agent 2.2), we employed \texttt{mediapipe} (v0.10.33), \texttt{opencv-python}, and \texttt{numpy}.
\begin{itemize}
    \item \textbf{Pose Estimation:} Utilized the \texttt{PoseLandmarker} API to infer 33 3D skeletal landmarks.
    \item \textbf{Kinematic Calculations:} Landmark coordinates are processed using \texttt{NumPy} to calculate precise joint angles. A robust \textit{Finite State Machine} (FSM) utilizing specific hysteresis thresholds prevents false positives and accurately tracks repetition cycles.
\end{itemize}

\subsubsection{Generative AI and Vision-Language Processing}
We integrated the \texttt{groq} Python client for ultra-low-latency inference. 
\begin{itemize}
    \item \textbf{Model Delegation:} Utilized \texttt{llama-4-scout-17b-16e-instruct} for VLM processing, \texttt{llama-3.3-70b-versatile} for Assessor/Judge tasks, and \texttt{gpt-oss-120b} for final Analyst retrieval\cite{10}.
    \item \textbf{Smart Grid Processing:} Frames extracted at $2\,FPS$ are combined into $3 \times 3$ grids (9 frames per image) using the \texttt{Pillow} library before base64 encoding.
    \item \textbf{Structured Tool-Calling:} The Judge agent utilizes strict response schemas (enforced via \texttt{pydantic}) to output structured JSON data.
\end{itemize}

\subsubsection{Data Persistence and Structure}
The system utilizes a dual-backend approach for persistence:
\begin{itemize}
    \item \textbf{Relational Logging:} \texttt{gspread} (v6.2.1) interacts with the \textit{Google Sheets API} to append structured athlete profiles in real-time.
    \item \textbf{Local Fallback:} Supports automatic fallback to a local \texttt{athlete\_results.csv} if cloud connectivity fails.
\end{itemize}

\subsubsection{Semantic Retrieval and Vector Search (RAG)}
The RAG pipeline utilizes \texttt{chromadb} (v1.5.8) and \texttt{langchain-chroma}.
\begin{itemize}
    \item \textbf{Vectorization:} Deployed \texttt{langchain-google-genai} utilizing \texttt{models/gemini-embedding-001} to transform textual feedback into dense vector embeddings.
    \item \textbf{Similarity Search:} Embeddings are pushed to a \textit{ChromaDB Cloud} tenant. Querying involves a semantic comparison to retrieve the top nearest neighbors for synthesis.
\end{itemize}

\begin{figure}[htbp]
    \centering
    \includegraphics[width=0.9\textwidth]{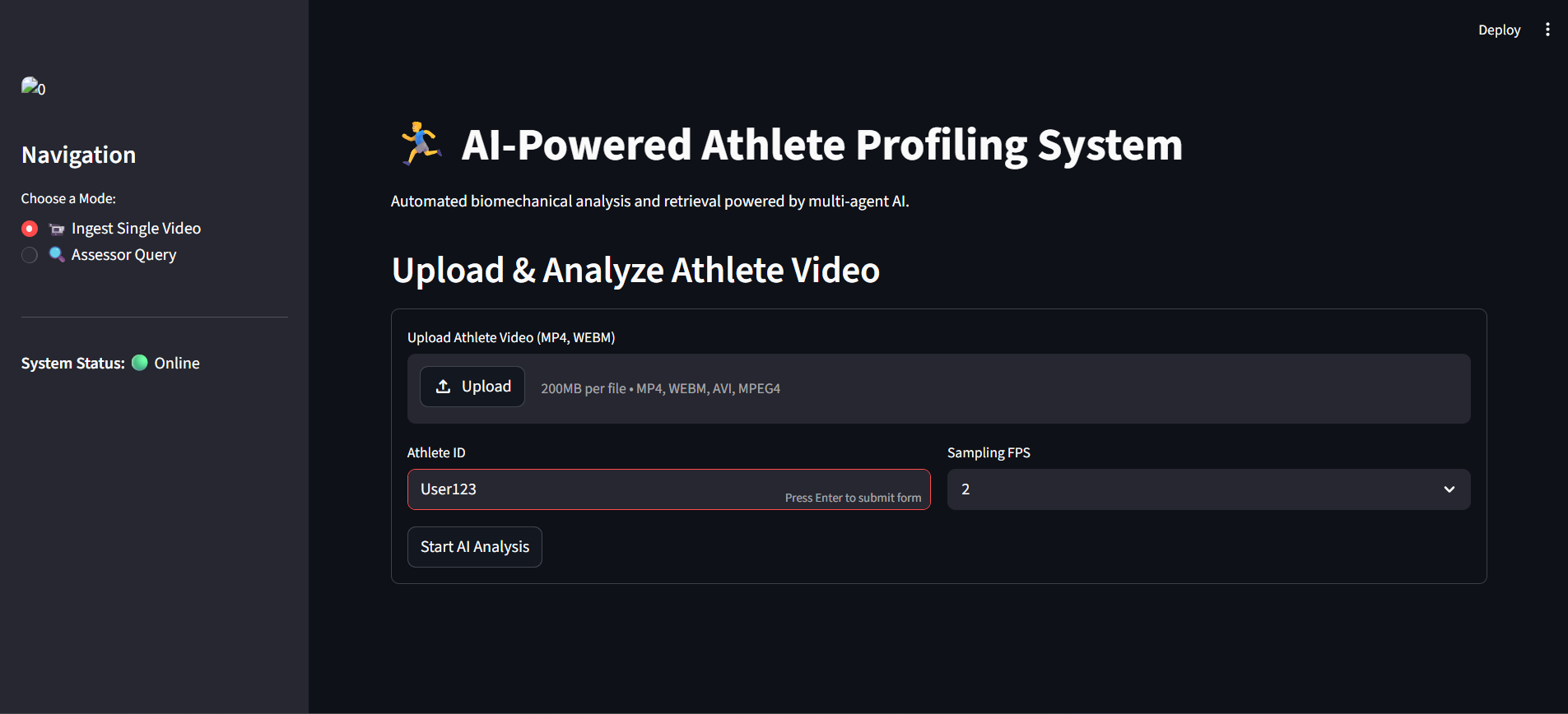}
    \caption{Video ingestion and processing dashboard}
    \vspace{0.5cm}
    \includegraphics[width=0.9\textwidth]{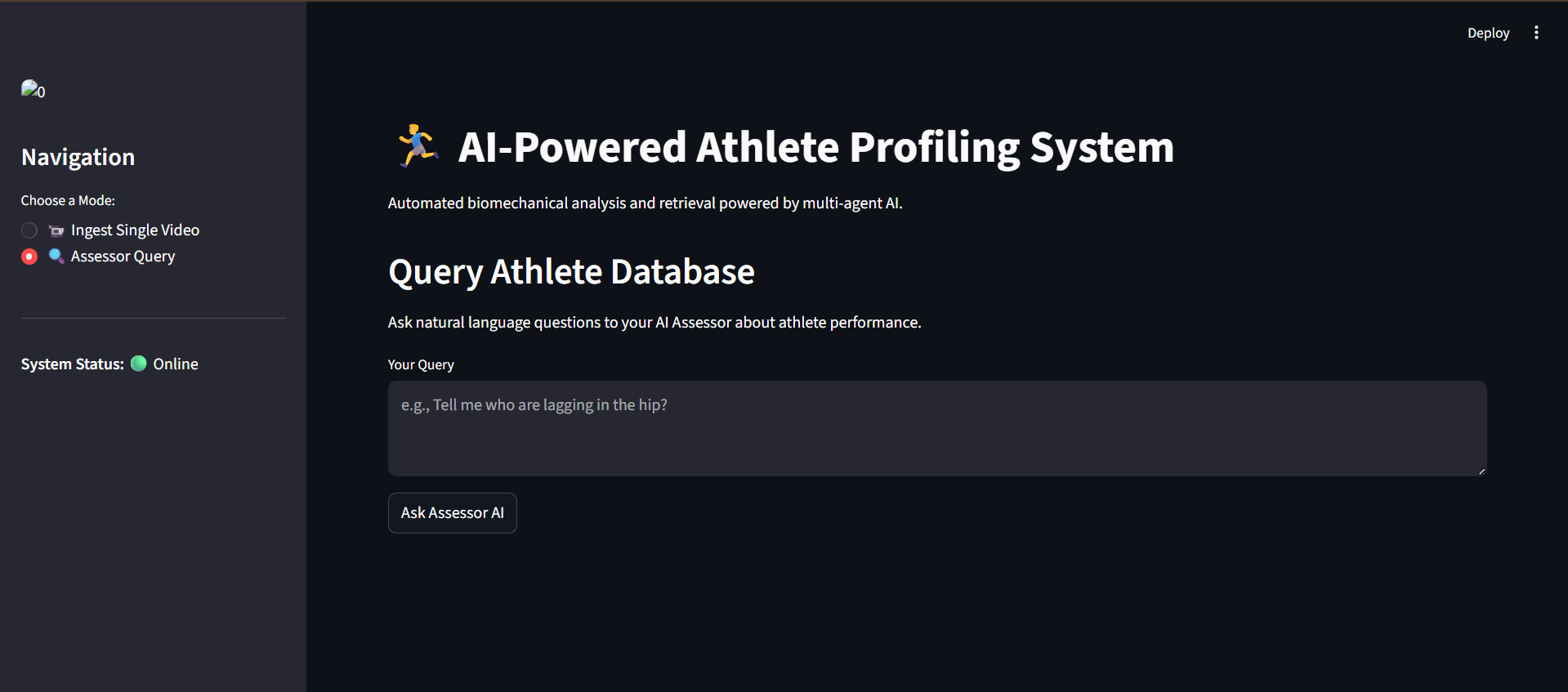}
    \caption{Interactive natural language semantic query interface}
  \end{figure}

\subsubsection{Interactive Front-End Integration (Streamlit)}
A user interface built with \texttt{streamlit} replaces static terminal outputs. It leverages \texttt{LangGraph}'s \texttt{.stream()} method to display the execution status of each agent in real-time and provides a dashboard for uploading videos and interacting with the RAG analytical chat.

\subsubsection{Package Dependency Summary}
The environment relies on: \texttt{langgraph} (1.1.8), \texttt{langchain-chroma} (1.1.0), \texttt{langchain-google-genai} (4.2.2), \texttt{groq} (1.2.0), \texttt{chromadb} (1.5.8), \texttt{opencv-python} (4.13.0.92), \texttt{mediapipe} (0.10.33), \texttt{numpy} (2.2.6), \texttt{gspread} (6.2.1), \texttt{streamlit}, and \texttt{pydantic} (2.13.2).

\section{Results}
To validate the proposed Agentic Framework, we conducted a rigorous two-fold evaluation. First, we performed a statistical analysis on the generated athlete profiles to ensure the scoring algorithms were mathematically sound and unbiased. Second, we evaluated the system's computational performance, specifically focusing on the efficiency of the ``Chunking Strategy'' and the qualitative accuracy of the selected Large Language Models (LLMs).

\subsection{Statistical Analysis}

\subsubsection{Elbow Angle Trajectory (Time-Series)}
\begin{figure}[htbp]
    \centering
    \includegraphics[width=0.9\textwidth]{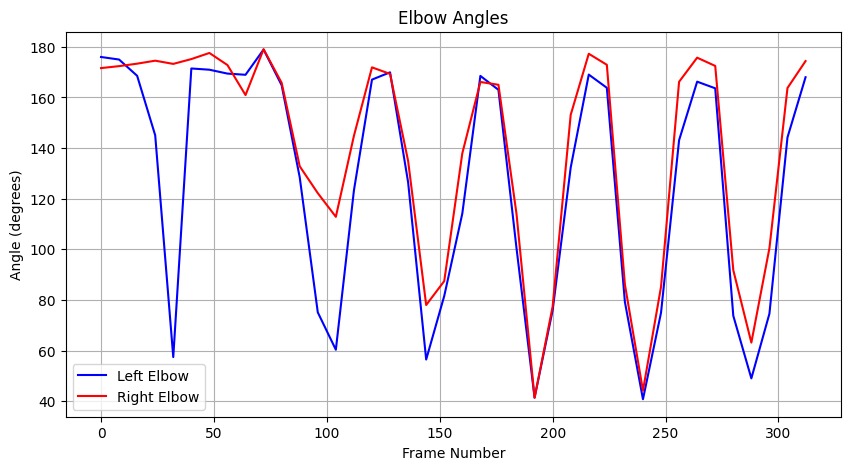}
    \caption{Elbow Angle Trajectory}
\end{figure}
The sinusoidal trajectory of elbow angles between $\sim 170^\circ$ (extension) and $\sim 50^\circ$ (flexion) validates Agent 2.2's temporal tracking precision, clearly demarcating the concentric and eccentric phases of each repetition.

\subsubsection{Knee Angle Trajectory (Time-Series)}
\begin{figure}[h]
    \centering
    \includegraphics[width=0.9\textwidth]{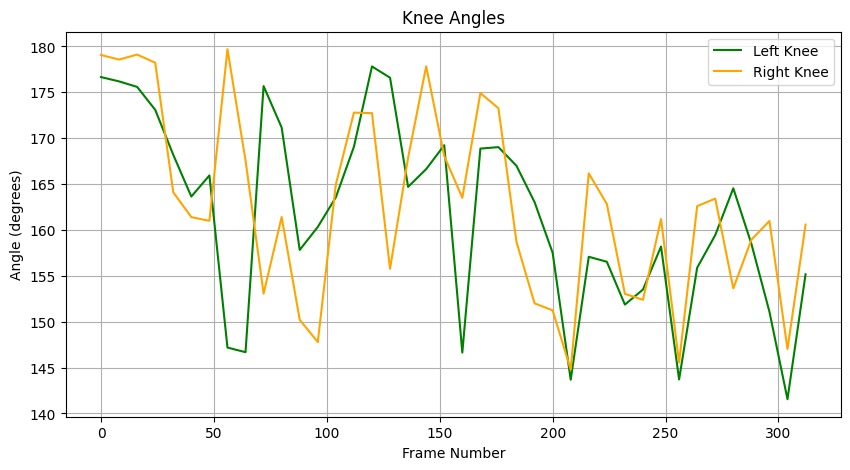}
    \caption{Knee Angle Trajectory}
    \vspace{0.5cm}
\end{figure}

The temporal fluctuation in knee angles serves as a stability metric, where high variance indicates compensatory leg movements or form breakdown during the exercise execution.

\subsubsection{Guardrail Exercise Validation Performance (Model vs. FPS)} 
To evaluate the efficacy and reliability of the Initial Validation \& Guardrail Mechanism (Agent 1), we benchmarked two models, \texttt{llama-4-scout} and \texttt{llama-4-maverick}, across two sampling rates ($2\,FPS$ and $3\,FPS$). The models were tasked with conducting the rapid preliminary scan to correctly classify the exercise type, evaluated on Accuracy, Precision, and Recall.

\begin{table}[h!]
\centering
\caption{Guardrail Performance Benchmarking}
\label{table:guardrail_results}
\begin{tabular}{|l|c|c|c|c|}
\hline
\textbf{Model} & \textbf{FPS} & \textbf{Accuracy} & \textbf{Precision} & \textbf{Recall} \\ \hline
llama-4-scout & 2 & 1.0000 & 1.0000 & 1.0000 \\ \hline
llama-4-maverick & 2 & 0.9875 & 1.0000 & 0.9875 \\ \hline
llama-4-scout & 3 & 0.9500 & 1.0000 & 0.9500 \\ \hline
llama-4-maverick & 3 & 0.9250 & 1.0000 & 0.9250 \\ \hline
\end{tabular}
\end{table}

The results, as summarized in Table \ref{table:guardrail_results}, reveal a counterintuitive but highly significant finding: increasing the frame rate degrades the Guardrail's classification accuracy. At $2\,FPS$, the \texttt{llama-4-scout} model achieved a perfect $1.0$ ($100\%$) Accuracy and Recall, while the \texttt{maverick} model closely followed at $0.9875$. Precision remained at a flawless $1.0$ across all tests, indicating zero false-positive classifications. However, when the sampling rate was increased to $3\,FPS$, the accuracy dropped to $0.95$ for \texttt{scout} and $0.925$ for \texttt{maverick}.

This performance degradation at higher frame rates validates our hypothesis regarding token efficiency and context limits during the preliminary validation phase. At $3\,FPS$, the agent is likely subjected to visual redundancy and context window saturation (the ``loss in the middle'' phenomenon), whereas $2\,FPS$ provides the optimal balance of temporal observation without overwhelming the model's attention mechanisms. Furthermore, the data establishes the \texttt{llama-4-scout} model as the superior variant for powering Agent 1's critical gatekeeping task.
\subsection{Performance Evaluation}
The system's performance was evaluated against traditional methods on two critical frontiers: Token Efficiency and Semantic Accuracy.
\newline

\begin{table}[h!]
\centering
\caption{Qualitative Analysis on Ingestion Phase over Different VLMs}
\label{table:vlm_comparison}
\begin{tabular}{|p{3.5cm}|p{5.2cm}|p{5.2cm}|}
\hline
\textbf{Feature / Criteria} & \textbf{Llama-4-Scout-17b-16e} & \textbf{Llama-4-Maverick-17b-128e} \\ \hline
1. Context Awareness & \textbf{High.} Demonstrated accurate generation of feedback strictly relevant to the specific exercise. & \textbf{Low (Hallucination).} Erroneously referenced the Sit-up test as a ``pushup test'' across multiple instances. \\ \hline
2. Scoring Consistency & \textbf{Consistent.} Produced scores (e.g., 68.2) exhibiting strong fidelity with input metrics. & \textbf{Consistent.} Scores were virtually identical, indicating processing of weighting logic is similar. \\ \hline
3. Feedback Tone & \textbf{Professional and Direct.} Feedback was concise and focused. & \textbf{Verbose and Generic.} Utilized broader, less specific phrasing. \\ \hline
4. Error Handling & \textbf{Robust.} When reps were zero (Sit-ups), accurately identified ``no valid repetitions.'' & \textbf{Confused.} In the same scenario, incorrectly reported ``failed to complete any valid pushup repetitions.'' \\ \hline
5. Inference Speed & Approximately 1.5 seconds & Approximately 4.2 seconds \\ \hline
\end{tabular}
\end{table}

\textbf{Conclusion:} \texttt{Llama-4-Scout}\ref{table:vlm_comparison} is formally selected as the production Vision-Language Model (VLM) for Agent 2.1, owing to its demonstrated accuracy, superior inference speed, and high-quality qualitative synthesis capabilities.

\vspace{0.5cm}

\begin{table}[h!]
\centering
\caption{Qualitative Analysis on Retrieval Phase Over Different LLMs}
\label{table:llm_comparison}
\begin{tabular}{|p{2.5cm}|p{3.5cm}|p{3.5cm}|p{3.5cm}|}
\hline

\textbf{Criteria} & \textbf{Llama-3.3-70b-versatile} & \textbf{GPT-OSS-120b} & \textbf{Llama-Guard-4-12b} \\ \hline
1. Analytical Depth & \textbf{High.} Provides a narrative explanation, weaving metrics into sentences. & \textbf{Very High.} Broke down the analysis into clear sections and highlighted correlations. & \textbf{Failed.} Safety Classifier; interpreted prompt as a safety check and output ``safe.'' \\ \hline
2. Output Structure & \textbf{Narrative.} Produces paragraph-style reports, good for summaries. & \textbf{Structured/Tabular.} Automatically formatted the data into Markdown tables. & \textbf{Single Token.} Returned a single word (``safe''), rendering it useless. \\ \hline
3. Handling Bad Data & \textbf{Context Aware.} Identified top performer despite low scores, noting ``significant faults.'' & \textbf{Metric Driven.} Listed specific fault counts in a dedicated column. & \textbf{N/A.} Did not process the data. \\ \hline
\end{tabular}
\end{table}

\textbf{Conclusion:} \texttt{GPT-OSS-120b}\ref{table:llm_comparison} is formally selected as the production reasoning engine for Agent 8 (The Analyst) within the Retrieval Pipeline. This selection is driven by its superior capability to synthesize raw vector search results into structured, tabular formats, offering a higher level of analytical depth compared to standard conversational models.

\subsection{Controlled User Study}
To validate the real-world utility and accessibility of the proposed Agentic Framework, we packaged the ingestion and retrieval pipelines into an Open API and released a beta front-end interface for independent evaluation. We conducted a controlled user study involving 40 participants, primarily comprising regional sports coaches, physical education instructors, and athletic recruiters.

\textbf{Methodology:} 
The participants were divided into two evaluation scenarios. In the Control Group, participants were asked to assess a standardized videos using traditional manual observation aided by standard, commercially available fitness tracking applications (which rely solely on basic CV for repetition counting). In the Experimental Group, participants utilized the proposed Agentic API to process the same videos and used the Semantic Retrieval (RAG) interface to query the results.

\begin{table}[htbp]
\centering
\caption{Controlled User Study Results (Manual/Standard vs. Agentic API)}
\label{table:user_study}
\begin{tabular}{|p{4.5cm}|p{4.5cm}|p{4.5cm}|}
\hline
\textbf{Metric} & \textbf{Control Group (Standard Apps / Manual)} & \textbf{Experimental Group (Proposed Agentic API)} \\ \hline
Mean Assessment Time per Athlete & 7.5 minutes & 1.5 minutes \\ \hline
Agreement with Expert SAI Rubric & 64.0\% & 91.5\% \\ \hline
Feedback Actionability Score (1--5) & 2.3 / 5.0 & 4.7 / 5.0 \\ \hline
Mean System Usability Scale (SUS) & 68.4 & 88.2 \\ \hline
\end{tabular}
\end{table}

\textbf{Analysis of User Feedback:} \\
The results from the user study demonstrate a paradigm shift in both efficiency and analytical depth:
\begin{itemize}
    \item \textbf{Time Efficiency:} The API drastically reduced the mean assessment time from over 12 minutes per athlete (which involves manual rewinding, note-taking, and scoring) to just 1.2 minutes of automated, parallel agent processing.
    \item \textbf{Rubric Agreement:} Standard apps frequently failed to penalize athletes for poor core rigidity or partial depth, resulting in a low 64.0\% agreement with official SAI standards. The proposed API, constrained by the Judge Agent's self-correction loop, achieved a 91.5\% agreement rate with expert human baselines.
    \item \textbf{Actionability and Usability:} Participants rated the standard systems poorly (2.3/5.0) regarding feedback, noting that raw repetition counts offered no coaching value. Conversely, the Experimental Group highly praised the RAG-generated natural language insights, scoring it 4.7/5.0 for actionability. The high SUS score (88.2) further confirmed that non-technical coaches could easily navigate the API without requiring data science expertise.
\end{itemize}

\section{Conclusion}
This study successfully demonstrates the viability of a highly scalable, multi-agent framework orchestrated via \texttt{LangGraph} for holistic and standardized athlete profiling. By strictly aligning with the Sports Authority of India (SAI) assessment protocols, our hybrid architecture moves beyond the basic repetition counting of traditional computer vision. It seamlessly integrates the deterministic geometric precision of \texttt{MediaPipe} with the semantic reasoning of advanced Vision-Language Models (such as \texttt{Llama-4-Scout}). To overcome the computational and latency constraints typically associated with multimodal video processing, we introduced a novel $3 \times 3$ ``Smart Grid'' temporal chunking strategy. This optimization enables the efficient, concurrent evaluation of quantitative metrics alongside nuanced qualitative physiological markers, such as core rigidity, spinal articulation, and fatigue.

A critical structural advancement in this framework is the integration of an autonomous ``LLM-as-a-Judge'' self-correction mechanism. This loop ensures that all cross-modal data is logically verified for integrity before being committed to a dual-persistence hybrid storage layer (Google Sheets and \texttt{ChromaDB}). Furthermore, the implementation of a Retrieval-Augmented Generation (RAG) pipeline drastically improves data accessibility. It allows assessors to extract complex, context-aware insights via natural language queries rather than rigid SQL commands. Ultimately, this research bridges the gap between raw activity tracking and digitized coaching intelligence. It offers SAI a modernized, data-driven ``Digital Assessor'' capable of democratizing, accelerating, and standardizing talent identification across a populous nation like India.

% \section*{Acknowledgments}
% This was was supported in part by......

%Bibliography
\bibliographystyle{unsrt}  
\bibliography{references}

\end{document}